# A Transfer Learning Framework for Proactive Ramp Metering Performance Assessment

**Xiaobo Ma, Ph.D.**
Department of Civil & Architectural Engineering & Mechanics
The University of Arizona
1209 E 2nd St, Tucson, AZ 85721
Email: xiaoboma@arizona.edu

**Adrian Cottam, Ph.D.**
Department of Civil and Architectural Engineering and Mechanics
University of Arizona
1209 E 2nd St, Tucson, AZ 85721
Email: acottam1@arizona.edu

**Mohammad Razaur Rahman Shaon, Ph.D.**
Connecticut Transportation Safety Research Center
University of Connecticut
270 Middle Turnpike, Storrs, CT 06269
Email: mrr.shaon@uconn.edu

**Yao-Jan Wu, Ph.D.**
Department of Civil and Architectural Engineering and Mechanics
The University of Arizona
1209 E 2nd St, Tucson, AZ 85721
Email: yaojan@arizona.edu

**ABSTRACT**

Transportation agencies need to assess ramp metering performance when deploying or expanding a ramp metering system. The evaluation of a ramp metering strategy is primarily centered around examining its impact on freeway traffic mobility. One way these effects can be explored is by comparing traffic states, such as the speed before and after the ramp metering strategy has been altered. Predicting freeway traffic states for the "after" scenarios following the implementation of a new ramp metering control strategy could offer valuable insights into the potential effectiveness of the target strategy. However, the use of machine learning methods in predicting the freeway traffic state for the "after" scenarios and evaluating the effectiveness of transportation policies or traffic control strategies such as ramp metering is somewhat limited in the current literature. To bridge the research gap, this study presents a framework for predicting freeway traffic parameters (speed, occupancy, and flow rate) for the "after" situations when a new ramp metering control strategy is implemented. By learning the association between the spatial-temporal features of traffic states in before and after situations for known freeway segments, the proposed framework can transfer this learning to predict the traffic parameters for new freeway segments. The proposed framework is built upon a transfer learning model. Experimental results show that the proposed framework is feasible for use as an alternative for predicting freeway traffic parameters to proactively evaluate ramp metering performance.

**Keywords:** Ramp metering evaluation, Transfer learning, Traffic prediction

# 1. INTRODUCTION

Ramp metering is an essential technique used by traffic professionals to alleviate congestion and promote efficiency on freeways. The principle behind ramp metering is to restrict the rate at which vehicles can enter the freeway, so that the freeway does not become congested. Different ramp metering strategies have been extensively investigated over the years, resulting in several exceptional metering strategies. One such strategy is Asservissement Linéaire d'Entrée Autoroutière (ALINEA), a fundamental local traffic-responsive feedback-based metering strategy (Hadj-Salem et al., 1990). With advancements in technology, adaptive ramp metering strategies, such as the Zone algorithm, Fuzzy logic algorithm, System-wide adaptive ramp metering (SWARM) algorithm, and Heuristic ramp-metering coordination (HERO) algorithm have been introduced (Shaaban et al., 2016). With all of these available ramp metering strategies, it has become paramount to measure the effectiveness of ramp metering control strategies. The evaluation of a ramp metering strategy is usually focused on exploring the effect the ramp metering strategy has on traffic mobility by comparing traffic states such as speed before and after the strategy is implemented on a freeway. Ramp metering performance evaluation plays an important role not only in designing a more efficient ramp metering control algorithm, but also in freeway management, operations, and planning(Luo et al., 2022; Ma, 2022; Ma et al., 2020).

In the past, traffic simulation has been widely used to evaluate the effectiveness of ramp metering strategies (Han et al., 2020). Simulation is considered a proactive evaluation approach, as ramp metering performance can be evaluated before field implementation. Currently, a properly calibrated traffic simulation model remains a popular and viable approach for ramp metering evaluation. However, even when the parameters are well-calibrated, simulation scenarios may still have deviations from real-world scenarios. Ke et al. (2021) calibrated and validated simulation scenarios using field data. Compared to the field data, the Mean Absolute Percent Error (MAPE) for calibrated flows and calibrated speeds were 9.2% and 11.2%, respectively (Ke et al., 2021). After the calibration process, Abdel-Aty et al. (2007) verified calibration results and found that the calibrated flows and speeds had average errors of 8.26% and 18.9%, respectively (Abdel-Aty et al., 2007). Deviations introduced in the simulation model calibration process may undermine the trustworthiness of the evaluation results produced by traffic simulation. Hence, simulating real-world traffic conditions by parameter calibration is still a demanding task for traffic simulation. In addition, assuming there is a well-built mathematical model that can accurately mimic the interaction between vehicles, environment, and infrastructures for a certain location. Considering different locations have different traffic patterns and the interaction between vehicles, environment, and infrastructures for each location is different from each other(Liu et al., 2021; Meng et al., 2016; Rubaiyat et al., 2018), it is impossible to transfer the well-built mathematical model to other locations as different locations have different traffic patterns. Hence, the transferability of models requires further investigation to improve their adaptability to various traffic patterns.

In practice, many factors influence the ramp metering performance, and data collected from the field is a realistic and reliable representation of actual traffic patterns. It's more meaningful and convincing to validate whether a ramp metering control strategy is able to improve traffic conditions based on the data collected from the field. As an alternative way of traffic simulation, a field operational test is a commonly used reactive approach for ramp metering performance evaluation, as the evaluation results are available only after the field implementation of the metering strategy. Transportation agencies have assessed various dimensions of ramp metering performance after deploying or expanding a ramp metering system by collecting field data (Bhouri et al., 2013). In a field operational test, the effects of ramp metering can be observed only when it is implemented in the field, causing the field operational test to be lengthy and time-consuming (Hasan et al., 2002). Similar to traffic simulation, far too little attention has been given to rescaling or reusing the knowledge learned from known freeway segments influenced by a specific ramp metering control strategy for predicting traffic mobility at new freeway segments likewise influenced by the same strategy. Collecting and mining the data from known freeway segments to assist with traffic mobility prediction for new study sites can be a viable alternative for evaluating ramp metering performance.

Traffic prediction is another way of forecasting traffic mobility, and the predicted traffic parameters can be used to assist ramp metering evaluation. In recent years, machine learning algorithms have been extensively applied to solve diverse problems(Hu et al., 2020; Wu et al., 2022). The recent progress in Artificial Intelligence (AI) and Machine Learning (ML) has empowered researchers and practitioners to utilize a wide array of machine learning techniques(D.-F. Liu et al., 2022; Y. Yang et al., 2023; Zhao et al., 2020). These techniques encompass K-nearest neighbor(Li et al., 2023), random forest(Qu and Hickey, 2022; Wang and Qu, 2022), support vector machine(Wu et al., 2023), artificial neural network(Kong et al., 2021), convolutional neural network(Sun et al., 2022; X. Yang et al., 2023; Yi and Qu, 2022; Zhan et al., 2022; Zhang et al., 2022), generative adversarial network(Huang et al., 2020; Huang and Chiu, 2020), reinforcement learning(Bautista-Montesano et al., 2022; J. Chen et al., 2022; Z. Chen et al., 2022; He et al., 2022; Ke et al., 2020; Liu et al., 2023; Mei et al., 2023; Zhou et al., 2022), curriculum learning(Dou et al., 2023), transfer learning(W. Liu et al., 2022), contrastive learning(Pokle et al., 2022; Shen et al., 2021), representation learning(Dou et al., 2022a, 2022b), incremental learning(Xiang and Shlizerman, 2022), limitation learning(Ruan and Di, 2022), transformer(Tian et al., 2023; Zhang and Zhou, 2023; Zhou et al., 2023), and natural language processing(Yan et al., 2023). The application of these machine learning techniques has gained significant popularity in solving various tasks, particularly in the transportation domain(Zhang and Lin, 2022). There is a considerable amount of literature which used machine learning techiniques for traffic parameter prediction. Long short-term memory recurrent neural network (Ma et al., 2015), convolutional neural network (Ma et al., 2017), and generative adversarial network (Zhang et al., 2021) have been introduced in traffic prediction with promising performance. To date, most published studies focused on short-term or long-term traffic predictions without considering the transferability of the developed methodology to new study sites. The utilization of machine learning methodologies in forecasting freeway traffic mobility parameters, such as speed, after implementation of a ramp metering control strategy has been rarely studied in previous literature.

To address the above-mentioned research gaps, this research study introduces a purely data-driven framework to proactively evaluate ramp metering performance by predicting speed, traffic flow, and occupancy for both on-ramps and mainlines in the after periods. The proposed framework aims to provide some insights on the impact of ramp metering on the freeway by associating spatial-temporal traffic features that derive from the before and after periods. Besides, this framework seeks to rescale or reuse the knowledge learned from known freeway segments for new study segment evaluation. Ridge regression (Hoerl and Kennard, 2000) and a Two-stage TrAdaBoost.R2 algorithm (Pardoe and Stone, 2010) are adopted in this framework to obtain the research objectives. When trying to interpret the impacts of ramp metering on the freeway, spatial-temporal traffic features are highly correlated with each other. In this situation, the coefficient estimates of the regression may change erratically in response to small changes in the model or the data (Du et al., 2022; Yao et al., 2020). Therefore, ridge regression is adopted in this study as it is most useful when there is multicollinearity in the features (Hoerl and Kennard, 2000). Besides, ridge regression can regularize coefficient estimates made by ordinary least squares to avoid over-fitting issues, causing the estimated coefficients to work better on new datasets (James et al., 2013). Before rescaling or reusing the knowledge learned from known freeway segments for new study segment evaluation, one thing that needs to be considered is that different freeway segments have different traffic patterns, distributions, and characteristics. Traffic prediction models need to be able to adapt to the change of traffic patterns and distributions to obtain satisfactory prediction results. To handle the distribution differences between known freeway segments and new freeway segments, a transfer learning method named Two-stage TrAdaBoost.R2 is introduced in this study. As an instance-based transfer learning method, the Two-stage TrAdaBoost.R2 is designed for solving regression problems (Luo and Paal, 2021). It relaxes the assumption that the data distributions of the source domain and target domain must be the same (Q Pan S.J., 2010), which leads to a positive effect on real-world applications when the source and target-domain data have different distributions (Tang et al., 2020; Yehia et al., 2021).

In this study, by interpreting the impact of a ramp metering control strategy on known freeway segments, a ridge regression can provide reference to the variable selection when conducting traffic prediction for new study segments that are influenced by the same ramp metering control strategy as the known segments. Based on the ridge regression results, the most influential variables are selected as inputs for the transfer learning model. To enhance the transferability and adaptability to new scenarios when performing ramp metering evaluation, Two-stage TrAdaBoost.R2 is used to predict traffic parameters for the "after" situations of new freeway segments. By comparing the predicted traffic parameters with the traffic parameters in the before periods, the impacts of the studied ramp metering control strategy on new freeway segments can be determined (e.g., increase or decrease in mainline traffic speed). By predicting traffic parameters in the "after" period, engineers could evaluate the feasibility of applying a new ramp metering control strategy to target on-ramps before the ramp metering control strategy is implemented in the field. The effectiveness of the proposed framework is examined by comparing the results with real-world traffic data collected from the southbound of State Route 51 (SR51 SB) of the Phoenix Metropolitan area. Experiment results show that the proposed framework can achieve superior prediction performance over common prediction methods. We should note here that this research doesn't intend to replace traffic simulation or field operational testing. Rather, we want to provide an alternative approach that can be utilized to predict ramp metering performance when real-world field data is available. Since each method has its own strengths and weaknesses, transportation agencies should carefully select the most suitable method that fits the specific application scenario and meets special needs.

The contributions of this study are summarized as follows:

- This study aims to employ a data-driven framework to provide insights on the impacts of ramp metering on the freeway by associating spatial-temporal traffic features that derive from before and after situations, which is a complement to existing literature that use mathematical equations to represent complex traffic dynamics.
- A ridge regression model is utilized to enhance the interpretability of the proposed method in explaining the influence of ramp metering on freeway mobility.
- Since each location has its own unique traffic patterns, it would not make sense to use a standardized mathematical model across the board. Instead, this study uses a transfer learning model to address the challenge of adapting to the traffic patterns of different locations.
- Compared with the state-of-the-art short-term or long-term traffic prediction methods, the proposed framework can predict traffic parameters accurately when a freeway is impacted by a change in ramp metering.
- This study collects and mines data from known freeway segments to assist with traffic prediction for new study freeway segments. It can rescale and reuse the knowledge learned from known freeway segments for new study site evaluation, which provides a viable approach that is different from traffic simulation and field operational testing for ramp metering evaluation.
- The proposed framework is demonstrated to have superior spatial-temporal transferability for traffic prediction of new freeway segments compared with other state-of-the-art regression methods.

The remainder of this paper is organized as follows: preliminaries and problem definition are described in the second section. The third section introduces the proposed framework and methodologies. In the fourth section, a case study is presented to examine the spatial-temporal transferability of the proposed framework. Lastly, section five provides the conclusion and future work.

## 2. PROBLEM DEFINITION AND PRELIMINARIES

The notations, traffic variable definitions, and data encoding are presented first. Then the temporal correction of traffic variables is defined. The problem addressed in this study is formulated at the end of this section.

### 2.1. Notations

For the convenience of modeling and problem definition, Table 1 lists key indices, sets, and parameters used. Figure 1 provides a freeway stretch which assumes a long freeway is divided into $R$ sections. Each section includes an upstream segment, a downstream segment, an on-ramp segment, and an off-ramp segment. The red rectangles indicate the locations of loop detectors. The tilde symbol $\tilde{}$ and caret symbol $\hat{}$ indicate the variables that come from the before and after situations of ramp metering control strategy change, respectively. In this study, traffic flow and occupancy were collected from loop detector data, and segment-based speed data was provided by probe vehicle-based data collected from INRIX. The data collected from INRIX provides segment-based speed and travel time of each segment for every minute. Referring to Figure 1, probe vehicle-based data from INRIX can provide segment-based speed data for upstream segments, on-ramp segments, and downstream segments.

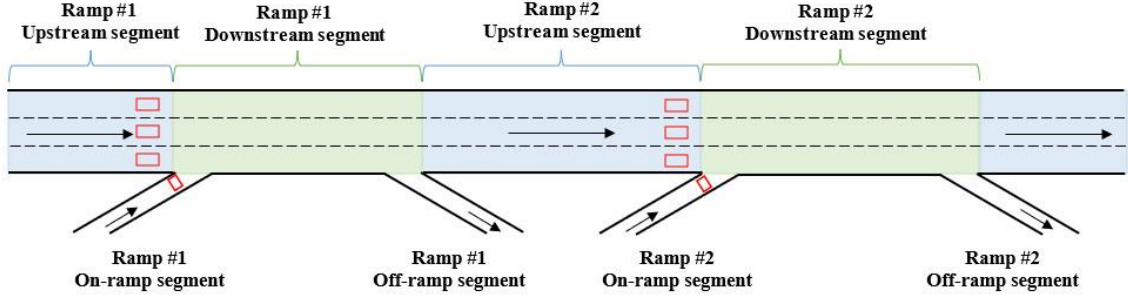

**Figure 1 Illustration of a Qualitative Freeway Stretch**

**Table 1 Notations and Parameters**

| Indices and sets | Definition |
|---|---|
| $r$ | Index of section |
| $u$ | Index of freeway segment; 1, if segment is upstream segment; 2, if segment is downstream segment; 3, if segment is on-ramp segment |
| $\tilde{t}$ | Index of the time interval for the before period |
| $\hat{t}$ | Index of the time interval for the after period |
| $\tilde{w}$ | Index of the week for the before period |
| $\hat{w}$ | Index of the week for the after period |
| $R$ | Set of sections |
| $U$ | Set of freeway segments, $U = \{1,2,3\}$ |
| $\tilde{T}$ | Set of time intervals for the before period |
| $\hat{T}$ | Set of time intervals for the after period |
| **Parameters** | **Definition** |
| $\widetilde{W}$ | Number of weeks for the before period |
| $\widehat{W}$ | Number of weeks for the after period |
| $v_{r,u}^{\widetilde{W},\tilde{t}}$ | Temporally corrected traffic speed of all $\widetilde{W}$ weeks in the $u$-th segment of the $r$-th section at the time interval $\tilde{t}$ |
| $v_{r,u}^{\widehat{W},\hat{t}}$ | Temporally corrected traffic speed of all $\widehat{W}$ weeks in the $u$-th segment of the $r$-th section at the time interval $\hat{t}$ |
| $o_{r,u}^{\widetilde{W},\tilde{t}}$ | Temporally corrected occupancy of all $\widetilde{W}$ weeks in the $u$-th segment of the $r$-th section at the time interval $\tilde{t}$ |
| $o_{r,u}^{\widehat{W},\hat{t}}$ | Temporally corrected occupancy of all $\widehat{W}$ weeks in the $u$-th segment of the $r$-th section at the time interval $\hat{t}$ |
| $q_{r,u}^{\widetilde{W},\tilde{t}}$ | Temporally corrected traffic flow rate of all $\widetilde{W}$ weeks in the $u$-th segment of the $r$-th section at the time interval $\tilde{t}$ |
| $q_{r,u}^{\widehat{W},\hat{t}}$ | Temporally corrected traffic flow rate of all $\widehat{W}$ weeks in the $u$-th segment of the $r$-th section at the time interval $\hat{t}$ |
| $\rho_{r,u}^{\widetilde{W},\tilde{t}}$ | Temporally corrected traffic density of all $\widetilde{W}$ weeks in the $u$-th segment of the $r$-th section at the time interval $\tilde{t}$ |
| $\tilde{\delta}^{\tilde{t}}$ | Attribute of minute-of-hour at the time interval $\tilde{t}$ |
| $\tilde{\lambda}^{\tilde{t}}$ | Attribute of hour-of-day at the time interval $\tilde{t}$ |
| $\widetilde{\varphi}^{\widetilde{t}}$ | Attribute of day-of-week at the time interval $\widetilde{t}$ |
| $\widetilde{y}_{r,u}^{\widetilde{w},\widetilde{t}}$ | Traffic variable reading collected from the $u$-th segment of the $r$-th section at the time interval $\widetilde{t}$ of $\widetilde{w}$-th week |
| $\widetilde{y}_{r,u}^{\widetilde{W},\widetilde{t}}$ | Average value over traffic variable readings collected from the $u$-th segment of the $r$-th section at the time interval $\widetilde{t}$ |
| $\widehat{y}_{r,u}^{\widehat{w},\widehat{t}}$ | Traffic variable reading collected from the $u$-th segment of the $r$-th section at the time interval $\widehat{t}$ of $\widehat{w}$-th week |
| $\widehat{y}_{r,u}^{\widehat{W},\widehat{t}}$ | Average value over traffic variable readings collected from the $u$-th segment of the $r$-th section at the time interval $\widehat{t}$ |

## 2.2. Variable definition and data encoding

The calculation process of the traffic variables listed in Table 1 is illustrated in this subsection. Let $V_n:\{v_1,\ldots,v_n\}$ denotes the time series of speed within $\widetilde{t}(\widehat{t})$ time interval.

**(1) Mean Speed**

Mean speed is the mean speed of all vehicles included in segment $u$ at the time interval $\widetilde{t}(\widehat{t})$. Mean speed is given by

$$\bar{v} = \frac{1}{n} * \sum_{l=1}^{n} v_l \quad (v_1 \leq v_l \leq v_n) \tag{1}$$

**(2) Occupancy**

Occupancy refers to the percentage of time that there is a vehicle over the detector in segment $u$ at the time interval $\widetilde{t}(\widehat{t})$.

**(3) Traffic Flow Rate**

Traffic flow rate (veh/hr/ln) is the number of vehicles leaving segment $u$ at the time interval $\widetilde{t}(\widehat{t})$, divided by the time interval $\widetilde{t}(\widehat{t})$ and lane numbers.

**(4) Traffic Density**

Traffic density (veh/ln/mi) is the number of vehicles in segment $u$ at the time interval $\widetilde{t}(\widehat{t})$, divided by segment length and lane numbers.

**(5) Minute-of-hour, hour-of-day, and day-of-week**

Each hour has four 15-minute intervals, the minute-of-hour (MOH) taking values from 1 to 4 to represent each interval. Hour-of-day (HOD) is represented by a number starting from 0 to 23. Day-of-week (DOW) is represented by a number ranging from 1 to 7, with Sunday encoded as 1 and Saturday encoded as 7.

## 2.3. Temporal Correction

The temporal correction method is widely used in the field of transportation to reduce performance evaluation biases (Yin et al., 2012). The temporal correction employed in this study is the calculation of mean values of time-of-week historical data. Let $\widetilde{W}$ and $\widehat{W}$ denote the number of weeks for the before and after periods. Then the temporal correction of traffic variables for before and after periods is calculated by Eq.(2) and Eq.(3), respectively.

$$\widetilde{y}_{r,u}^{\widetilde{W},\widetilde{t}} = \frac{\sum_{\widetilde{w}=1}^{\widetilde{W}} \widetilde{y}_{r,u}^{\widetilde{w},\widetilde{t}}}{\widetilde{W}}, \quad \widetilde{t} \in \widetilde{T}, \ r \in R, \ u \in U \tag{2}$$

$$\widehat{y}_{r,u}^{\widehat{W},\widehat{t}} = \frac{\sum_{\widehat{w}=1}^{\widehat{W}} \widehat{y}_{r,u}^{\widehat{w},\widehat{t}}}{\widehat{W}}, \quad \widehat{t} \in \widehat{T}, \ r \in R, \ u \in U \tag{3}$$

Note that the time interval $\widetilde{t}$ and $\widehat{t}$ are specified by identical minute-of-hour, hour-of-day, and day-of-week, $\widetilde{t}$ and $\widehat{t}$ are set as 15 minutes. $\widetilde{y}_{r,u}^{\widetilde{W},\widetilde{t}} \in \{v_{r,u}^{\widetilde{W},\widetilde{t}}, o_{r,u}^{\widetilde{W},\widetilde{t}}, q_{r,u}^{\widetilde{W},\widetilde{t}}, \rho_{r,u}^{\widetilde{W},\widetilde{t}}\}$, $\widehat{y}_{r,u}^{\widehat{W},\widehat{t}} \in \{v_{r,u}^{\widehat{W},\widehat{t}}, o_{r,u}^{\widehat{W},\widehat{t}}, q_{r,u}^{\widehat{W},\widehat{t}}\}$.

## 2.4. Problem formulation

From the perspective of ramp metering evaluation, currently, INRIX data and loop detector data are the most widely used data sources for ramp metering evaluation by federal, state, and local agencies. In addition, traffic volume, speed, and occupancy that are calculated based on INRIX data and loop detector data are three major performance measures adopted for ramp metering evaluation in practice. It is for these reasons this study seeks to predict traffic volumes, speeds, and occupancies for the "after" situations where a new ramp metering control strategy is implemented. Moreover, by comparing predicted traffic parameters with those in the pre-implementation period, the impact of ramp metering control strategies can be evaluated. This study tries to predict speed, occupancy, and flow rate for freeway segments through learning the impact of a ramp metering control strategy on known freeway segments and transferring the knowledge learned to new freeway segments that are likewise influenced by the same control strategy. Specifically, for the known freeway segments, a series of spatial-temporal features are proposed to describe the traffic states before and after the placement of a ramp metering control strategy. Then temporal corrected spatial-temporal traffic features of before and after situations are mapped in chronological order based on time-of-week. Learning a function $F(\cdot)$ by associating temporal corrected spatial-temporal traffic features between the before and after situations for known freeway segments, the learned function $F(\cdot)$ can be used for predicting speed, occupancy, and flow rate when the ramp metering control strategy is implemented on new freeway segments. Eq.(4), Eq.(5), and Eq.(6) show the functions for speed, occupancy, and flow rate prediction, respectively.

$$F_1\left(\left[v_{r,u}^{\widetilde{W},\tilde{t}}, o_{r,u}^{\widetilde{W},\tilde{t}}, q_{r,u}^{\widetilde{W},\tilde{t}}, \rho_{r,u}^{\widetilde{W},\tilde{t}}, \tilde{\delta}^{\tilde{t}}, \tilde{\lambda}^{\tilde{t}}, \tilde{\varphi}^{\tilde{t}}\right]\right) = \left[v_{r,u}^{\widehat{W},\hat{t}}\right], \tilde{t} \in \tilde{T}, \hat{t} \in \hat{T}, r \in R, u \in U \quad (4)$$

$$F_2\left(\left[v_{r,u}^{\widetilde{W},\tilde{t}}, o_{r,u}^{\widetilde{W},\tilde{t}}, q_{r,u}^{\widetilde{W},\tilde{t}}, \rho_{r,u}^{\widetilde{W},\tilde{t}}, \tilde{\delta}^{\tilde{t}}, \tilde{\lambda}^{\tilde{t}}, \tilde{\varphi}^{\tilde{t}}\right]\right) = \left[o_{r,u}^{\widehat{W},\hat{t}}\right], \tilde{t} \in \tilde{T}, \hat{t} \in \hat{T}, r \in R, u \in U \quad (5)$$

$$F_3\left(\left[v_{r,u}^{\widetilde{W},\tilde{t}}, o_{r,u}^{\widetilde{W},\tilde{t}}, q_{r,u}^{\widetilde{W},\tilde{t}}, \rho_{r,u}^{\widetilde{W},\tilde{t}}, \tilde{\delta}^{\tilde{t}}, \tilde{\lambda}^{\tilde{t}}, \tilde{\varphi}^{\tilde{t}}\right]\right) = \left[q_{r,u}^{\widehat{W},\hat{t}}\right], \tilde{t} \in \tilde{T}, \hat{t} \in \hat{T}, r \in R, u \in U \quad (6)$$

## 3. METHODOLOGY
### 3.1. Research framework

As shown in Figure 2, the main idea behind the proposed framework is the use of freeway data collected from the before and after scenarios of known freeway segments for prediction model training. As long as data collected from the before situation for new freeway segments are provided, the well-trained model can be used to predict the traffic speed, occupancy, and flow rate for the after situation of new freeway segments. The proposed framework in this study contains several steps for prediction. In the first step, spatial-temporal traffic features for before and after situations of known freeway segments are extracted. In the second step, the average values of the above-mentioned spatial-temporal traffic features are calculated by time-of-week for before and after situations. We define the second step as the temporal correction. Next, temporally corrected spatial-temporal traffic features of the before and after situations are mapped in chronological order based on time-of-week, and a ridge regression model is built to interpret the impacts of ramp metering control strategy changes on speed, occupancy, and flow rate by associating temporal corrected spatial-temporal traffic features that are derived from before and after situations. This ridge regression model is also used for selecting the most influential variables as the inputs for the transfer learning model. Lastly, the Two-stage

TrAdaBoost.R2 is employed to predict speed, occupancy, and flow rate for new freeway segments. When collecting data for new freeway segments, the spatial-temporal traffic features are selected exactly the same as the influential variables concluded from known freeway segments.

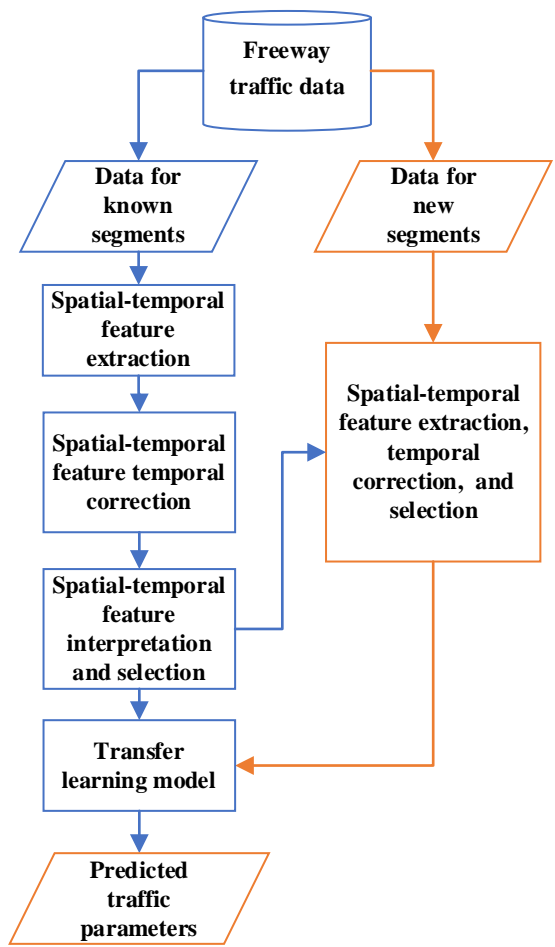

Figure 2 Framework for Traffic Parameter Prediction

## 3.2. Ridge regression

This study associates spatial-temporal traffic features from before and after situations by using ridge regression. Ridge regression is a supplement to least squares regression that trades unbiasedness in exchange for high numerical stability, thereby obtaining higher calculation accuracy (Hoerl and Kennard, 2000). The least square estimator may provide a good fit to the training data, but it will not fit sufficiently well to the test data. One way out of this situation is to abandon the requirement of an unbiased estimator. Assuming $X$ is a $n$ by $p$ matrix with centered columns, $Y$ is a centered $n$ vector.

The potential instability in the least square estimator
$$\beta = (X^T X)^{-1} X^T Y \tag{7}$$
could be improved by adding a small constant value $\lambda$ to the diagonal entries of the matrix $X^T X$ before taking its inverse. The result is the ridge regression estimator
$$\beta_{ridge} = (X^T X + \lambda I)^{-1} X^T Y \tag{8}$$

Where $I$ is the identity matrix, $\lambda$ is selected to give the minimum mean squared error.

Ridge regression places a particular form of constraint on the parameter: $\beta_{ridge}$ is chosen to minimize the penalized sum of squares.

$$\sum_{i=1}^{n}(y_i - \sum_{j=1}^{p} x_{ij}\beta_j)^2 + \lambda \sum_{j=1}^{p} \beta_j^2 \tag{9}$$

Which is equivalent to the minimization of $\sum_{i=1}^{n}(y_i - \sum_{j=1}^{p} x_{ij}\beta_j)^2$ subject to, for some $c > 0$, $\sum_{j=1}^{p} \beta_j^2 < c$, i.e. constraining the sum of the squared coefficients.

### 3.3. Two-stage TrAdaBoost.R2

Given source domain data $D_S = \{(x_{S_1}, y_{S_1}), \ldots, (x_{S_n}, y_{S_n})\}$, where $x_{S_i} \in \mathcal{X}_S$ is the data instance and $y_{S_i} \in \mathcal{Y}_S$ is the corresponding label. Target domain data is denoted as $D_T = \{(x_{T_1}, y_{T_1}), \ldots, (x_{T_k}, y_{T_k})\}$, where $x_{T_i} \in \mathcal{X}_T$ is the data instance and $y_{T_i} \in \mathcal{Y}_T$ is the corresponding label. In this study, $x_{S_i}$ and $x_{T_i}$ can be regarded as the spatial-temporal features of traffic states of before situations for known and new freeway segments, $y_{S_i}$ and $y_{T_i}$ represent traffic parameters of after situations for known and new freeway segments. When applying Two-stage TrAdaBoost.R2 to induce a predictive model, some labeled data in the target domain are required. To meet this requirement, $D_S' = \{(x_{S_j}, y_{S_j})\}$ ($D_S' \subsetneq D_S, 1 \leq j \leq m < n$) was created to act as a substitute for the labeled data in the target domain. $D_S'$ only contains pairs $(x_{S_j}, y_{S_j})$ in which the cosine similarity of $x_{S_j}(1 \leq j \leq m < n)$ and $x_{T_i}(1 \leq i \leq k)$ is less than or equal to a threshold value $\vartheta$. $\vartheta$ is set as a hyper-parameter.

As one of the instances-based inductive transfer learning methods, Two-stage TrAdaBoost.R2 is a boosting-based algorithm that aims to increase the prediction performance by combining weak estimators linearly to form a stronger one (Yehia et al., 2021).

$$F(x) = \sum_{t=1}^{T} \beta_t G(x, \gamma_t) \tag{10}$$

Here $\beta_t$ is the weight of the weak estimator, $G(x, \gamma_t)$ is the weak estimator, and $\gamma_t$ is the optimal parameter of the weak estimator.

Let $D_S$ (of size $n$) denotes labeled source domain data and $D_S'$ (of size $m$) denotes labeled data in the target domain. $D$ is the combination of $D_S$ and $D_S'$. $S$ is the number of steps, $F$ is the number of folds for cross-validation. $S$ and $F$ are set as 10 and 5, respectively. The algorithm of Two-stage TrAdaBoost.R2 is shown as follows:

**Input** $D$, $S$, and $F$. Setting the initial weight vector $\mathbf{w}^1$ such that $\omega_i^1 = \frac{1}{n+m}$ for $1 \leq i \leq n+m$

**For** $t = 1, \ldots, S$:
1. Call AdaBoost.R2 with $D$, estimator $G(x)$, and weight vector $\mathbf{w}^t$. $D_S$ is unchanged in this procedure. Obtaining an estimate $error_t$ of $model_t$ using $F$-fold cross-validation.
2. Call estimator $G(x)$ with $D$ and weight vector $\mathbf{w}^t$
3. Calculate the adjusted error $e_i^t$ for each instance as in AdaBoost.R2
4. Update the weight vector

$$\omega_i^{t+1} = \begin{cases} \dfrac{\omega_i^t \beta_t^{e_i^t}}{Z_t}, & 1 \leq i \leq n \\ \dfrac{\omega_i^t}{Z_t}, & n+1 \leq i \leq n+m \end{cases}$$

Where $Z_t$ is a normalizing constant, and $\beta_t$ is chosen such that the weight of the target (final $m$) is $\dfrac{m}{n+m} + \dfrac{t}{(S-1)}(1 - \dfrac{m}{n+m})$

**Output** $F(x) = model_t$, where $t = \operatorname{argmin}_i error_i$

## 4. CASE STUDY
### 4.1. Data description

The Arizona Department of Transportation (ADOT) uses adaptive ramp metering to actively control the traffic on the southbound direction of State Route 51 (SR51 SB) in the Phoenix Metropolitan area. ADOT changed its ramp metering control strategy from responsive ramp metering to adaptive ramp metering along SR51 SB, excepting Shea Blvd, McDowell Rd, and Union Hills Dr, on May 13, 2019. Shea Blvd's adaptive ramp metering came online on August 8, 2019. McDowell Rd and Union Hills Dr started adaptive ramp metering on November 1, 2019. Responsive control relies on loop detectors or other forms of traffic surveillance to select metering rates. Based on traffic conditions present on the ramp and at adjacent mainline locations, a responsive control strategy selects proper metering rates to remedy isolated congestion or safety-related problems. Responsive control cannot factor in conditions at adjacent ramps or throughout the freeway mainline. Adaptive control is responsive to both local and corridor-wide real-time traffic conditions. When calculating a metering rate, adaptive control takes into account traffic conditions upstream and downstream from an individual ramp.

Figure 3 shows the layout of the study corridor. In total, 14 on-ramps along SR51 SB were selected in this study. Table 2 lists segment lengths and lane numbers for each freeway segment. For convenience, each detector site is represented by its corresponding loop detector numbering. Considering that HOV lanes have lower traffic volumes and higher speeds than the general-purpose lanes, thus HOV lanes were excluded from this study. During the data collection period for this study, there was no inclement weather, significant incidents, bottlenecks, or severe data quality issues that could potentially affect the evaluation results.

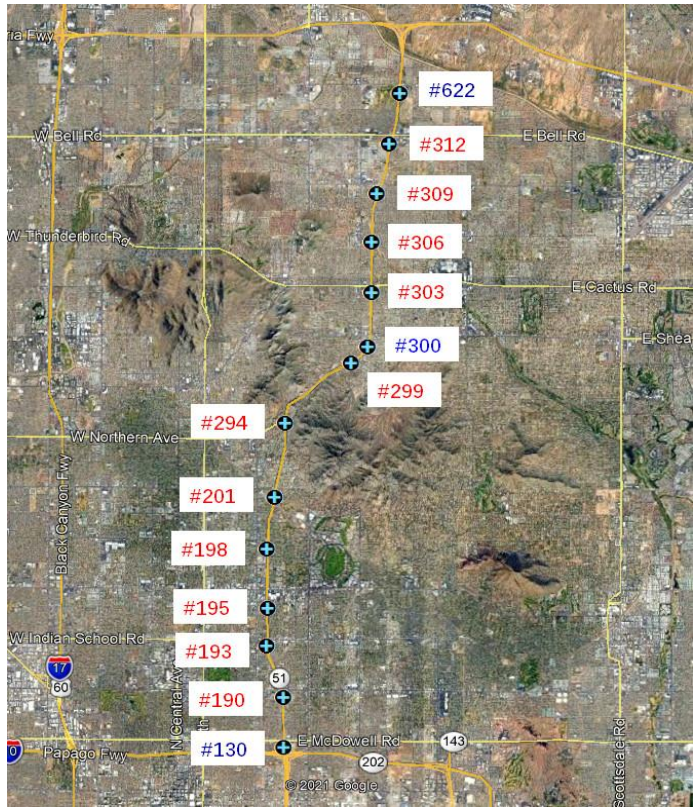
**Figure 3 Layout of the Study Corridor**

**Table 2 Segment Lengths and Lane Numbers for Freeway Segments**

| Name | Loop detector numbering | Upstream Segment Length (Miles) | Lane number | On-ramp Segment Length (Miles) | Lane number | Downstream Segment Length (Miles) | Lane number |
|---|---|---|---|---|---|---|---|
| MCDOWELL RD | #130 | 0.15 | 2 | 0.18 | 1 | 0.20 | 2 |
| THOMAS RD | #190 | 0.82 | 3 | 0.21 | 2 | 0.28 | 4 |
| INDIAN SCHOOL RD | #193 | 0.59 | 3 | 0.25 | 2 | 0.37 | 3 |
| HIGHLAND AVE | #195 | 0.55 | 3 | 0.26 | 2 | 0.19 | 4 |
| BETHANY HOME RD | #198 | 0.63 | 3 | 0.31 | 1 | 0.55 | 4 |
| GLENDALE AVE | #201 | 0.67 | 3 | 0.35 | 2 | 0.36 | 3 |
| NORTHERN AVE | #294 | 0.72 | 4 | 0.28 | 1 | 0.80 | 4 |
| W OF 32ND ST | #299 | 0.43 | 4 | 0.45 | 2 | 0.66 | 5 |
| SHEA BLVD | #300 | 0.87 | 3 | 0.38 | 2 | 0.43 | 4 |
| CACTUS RD | #303 | 0.66 | 3 | 0.26 | 2 | 0.36 | 4 |
| THUNDERBI-RD RD | #306 | 0.53 | 3 | 0.38 | 2 | 0.26 | 4 |
| GREENWAY RD | #309 | 0.68 | 3 | 0.34 | 2 | 0.53 | 3 |
| BELL RD | #312 | 0.72 | 3 | 0.26 | 2 | 0.27 | 4 |
| UNION HILLS DR | #622 | 0.70 | 4 | 0.34 | 2 | 0.29 | 4 |

The data used in this study were sourced from both loop detector data and probe vehicle data, both of which were provided by ADOT.
1) Loop detector-based data: ADOT had installed dual loop detectors on the mainline and ramps on the freeway in the Phoenix metropolitan area. At each detector site, traffic volume, vehicle speed, and occupancy are collected at 20-second intervals (Luo et al., 2022).
2) Probe vehicle-based data: The probe vehicle data used in this paper is INRIX data which provides segment-based speed and travel times of each segment every minute. INRIX collects data from a variety of sources. INRIX speed data includes speed data from probe-based systems - either agency-owned (Bluetooth) or third-party supplied (HERE Technologies, INRIX, TomTom.), and travel time is a derivative of speed data. Travel time also can be directly measured by probes, such as license plate recognition, toll tag transponders, Global Positioning Systems, and cell phone tracking. Alternatively, it can be estimated and predicted from other data sources.

Both data sources can provide speed data, however, the difference is that probe technology calculates speed as the average speed of vehicles over a length of road which is called space mean speed (SMS). Time mean speed (TMS) which is the arithmetic mean of vehicles' speed passing a point is the calculated speed for loop sensors. In this study, traffic flow and occupancy were collected from loop detector data and speed data was provided by probe vehicle-based data. Figure 4 and Figure 5 show screenshots of loop detector data and INRIX data samples respectively. In Figure 4, for each lane (SlotNumber) of each detector site (DetectorstationID), traffic volume, speed (unit: mph), and occupancy (unit: %) are provided. The loop detectors report measures (volume, speed, and occupancy) at 20-second intervals. Dual-loop detectors are installed on the mainlines and on-ramps of freeways.

| 1 | ID | TimeStamp | DetectorstationID | SlotNumber | Volume | Speed | Occupancy |
|---|---|---|---|---|---|---|---|
| 2 | 853115071 | 6/11/2019 6:00 | 312 | 1 | 4 | 0 | 8 |
| 3 | 853115071 | 6/11/2019 6:00 | 312 | 2 | 0 | 0 | 0 |
| 4 | 853115071 | 6/11/2019 6:00 | 312 | 9 | 4 | 0 | 4 |
| 5 | 853115071 | 6/11/2019 6:00 | 312 | 10 | 0 | 0 | 0 |
| 6 | 853115071 | 6/11/2019 6:00 | 312 | 33 | 8 | 66 | 12 |
| 7 | 853115071 | 6/11/2019 6:00 | 312 | 34 | 8 | 66 | 12 |
| 8 | 853115071 | 6/11/2019 6:00 | 312 | 35 | 7 | 76 | 8 |
| 9 | 853115071 | 6/11/2019 6:00 | 312 | 36 | 7 | 76 | 8 |
| 10 | 853115071 | 6/11/2019 6:00 | 312 | 37 | 8 | 80 | 8 |
| 11 | 853115071 | 6/11/2019 6:00 | 312 | 38 | 8 | 80 | 8 |

**Figure 4 Screenshot of Loop Detector Data Sample**

Figure 5 presents the details of the probe vehicle data. Probe vehicle data provides the segment-based speed and travel time of each segment every minute. Data from "speed" (unit: mph) and "travelTimeMinutes" (unit: minute) columns for each road segment (SegmentID) were used in this study.

| 1 | timestamp | SegmentID | type | speed | average | reference | score | confidenceValue | travelTimeMinutes |
|---|---|---|---|---|---|---|---|---|---|
| 2 | 4/16/2019 6:07 | 1226240265 | XDS | 50 | 63 | 63 | 30 | 32 | 0.49 |
| 3 | 4/16/2019 6:08 | 1226240265 | XDS | 52 | 63 | 63 | 30 | 44 | 0.47 |
| 4 | 4/16/2019 6:09 | 1226240265 | XDS | 52 | 63 | 63 | 30 | 55 | 0.47 |
| 5 | 4/16/2019 6:11 | 1226240265 | XDS | 52 | 63 | 63 | 30 | 57 | 0.47 |
| 6 | 4/16/2019 6:12 | 1226240265 | XDS | 61 | 63 | 63 | 30 | 99 | 0.41 |
| 7 | 4/16/2019 6:14 | 1226240265 | XDS | 60 | 63 | 63 | 30 | 100 | 0.41 |
| 8 | 4/16/2019 6:15 | 1226240265 | XDS | 57 | 63 | 63 | 30 | 92 | 0.44 |
| 9 | 4/16/2019 6:16 | 1226240265 | XDS | 53 | 63 | 63 | 30 | 79 | 0.46 |
| 10 | 4/16/2019 6:18 | 1226240265 | XDS | 54 | 63 | 63 | 30 | 79 | 0.46 |
| 11 | 4/16/2019 6:19 | 1226240265 | XDS | 58 | 63 | 63 | 30 | 99 | 0.42 |

**Figure 5 Screenshot of INRIX Data Sample**

Figure 6 visualizes traffic flow rate distributions for all the downstream segments. It can be seen that traffic flow rate distributions are quite similar among #294, #299, #300, #303, #306, and #309. But for other segments, their flow rate distributions are quite different from each other. It is worthwhile noting that the downstream segments of #130, #190, #193, #195, #198, and #201 suffer from recurrent traffic congestion during peak hours.

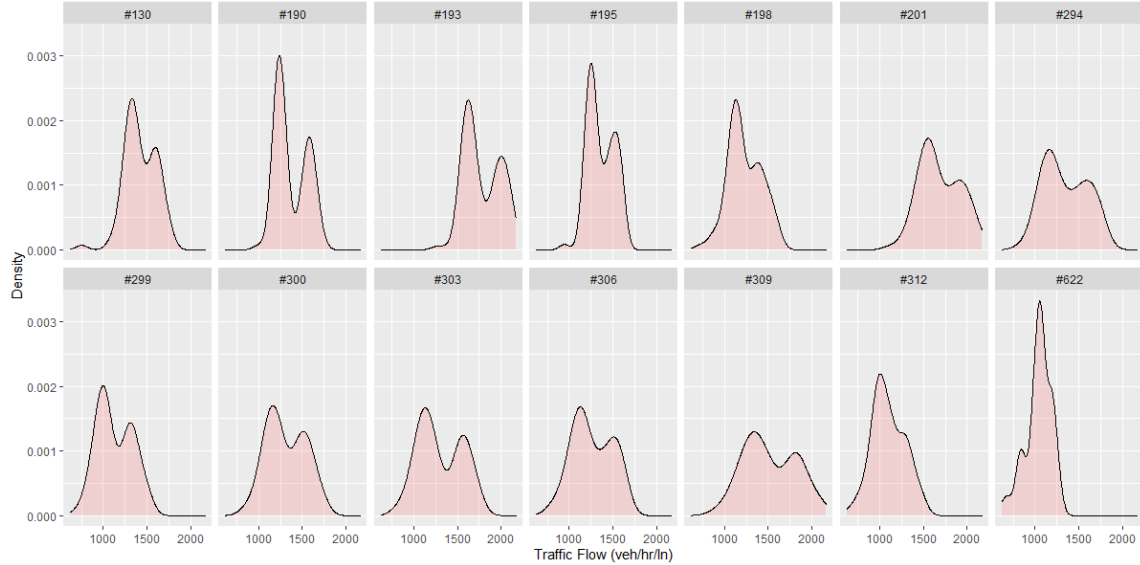

**Figure 6 Downstream Traffic Flow Distribution for Each Segment**

## 4.2. Experiment design
### 4.2.1. Data sample

To validate the proposed framework, this study chose one month of data before and one month of data after the changing of the ramp metering control strategy. It should be noted that when drivers encounter a ramp metering control strategy change, they may not know how to react and may exhibit unusual behaviors that could cause bias in the data being collected (Ma et al., 2020). In order to remove the influence of unusual driving behaviors coming from drivers, 1 month of data was eliminated after the ramp metering control strategy changed. For both before and after scenarios, only data for Tuesdays, Wednesdays, and Thursdays were used. This study only considers the operating times of ramp metering, which is in operation from 6 am - 9 am and 3 pm -7 pm. Table 3 presents the data collection duration for before and after periods of each on-ramp.

**Table 3 Before and After Data Selected for Each On-Ramp**

| Loop detector numbering | Before Period | After Period |
| --- | --- | --- |
| #130, #622 | 2019-10-06 ~ 2019-10-31 | 2019-12-01 ~ 2019-12-28 |
| #300 | 2019-07-11 ~ 2019-08-07 | 2019-09-05 ~ 2019-10-02 |
| #190,#193,#195,#198,#201,#294, #299,#303,#306,#309,#312 | 2019-04-14 ~ 2019-05-11 | 2019-06-09 ~ 2019-07-06 |

In the traffic variable calculation listed in section 2.2, probe vehicle data with a 1-min aggregation level was used to calculate any speed-related variables. The traffic volume data collected from the loop detectors had an aggregation level of 20 seconds and was summed up into 15-minute intervals.

The main intention of this study is to develop a more generalized model which can perform evaluation well on new freeway segments. To validate the generalization of the proposed framework, k-fold cross-validation was used. It is one of the most popular strategies widely used by data-driven research. In addition, it is a data partitioning strategy so that the dataset can be effectively used to build a more generalized model (Chen et al., 2020; Duan et al., 2022). In this study, 14 on-ramps along SR51 SB were selected, which leads to 14 sections available for k-fold cross-validation. Based on the previous description, each section includes an upstream segment, a downstream segment, an on-ramp segment, and an off-ramp segment. In the model training process, 13 sections were used for the model training and one section was used for performance evaluation. This procedure was repeated 14 times (iterations) so that 14 performance estimates (e.g. MAPE) for each iteration were obtained.

### 4.2.2. Model structure and hyper-parameters

Four representative machine learning algorithms, k-nearest neighbor (KNN), support vector regression (SVR), artificial neural network (ANN), and extreme gradient boosting (XGB), were selected as baseline models to examine the feasibility of the proposed framework for predicting traffic parameters in after situations of new freeway segments. These four state-of-the-art machine learning methods have been widely adopted as benchmark methods in terms of traffic prediction (Shaygan et al., 2022; Zhang et al., 2019). The proposed Two-stage TrAdaBoost.R2 is abbreviated to TRA in the following paragraphs for conciseness. Considering the values of algorithms' hyperparameters can greatly influence prediction accuracy. To select optimal values of hyperparameters for each algorithm, a grid search approach was developed. Table 4 shows grids of hyperparameters for machine learning algorithms. After conducting a grid search for the optimal structure, the best architectures for different tasks are listed in Table 5.

**Table 4 The Grids of Hyperparameters for Machine Learning Algorithms**

| Model | Hyperparameters |
|---|---|
| KNN | n_neighbors={1,3,5,7,9,11,13,15,17,19} |
| SVR | C={0.1, 1, 10, 100, 1000} <br> Gamma={1, 0.1, 0.01, 0.001, 0.0001} |
| ANN | hidden_layer_sizes={(50,100),(50,150),(50,200),(50,250),(50,300), (100,100),(100,150),(100,200),(100,250),(100,300)} |
| XGB | learning_rate={0.0001,0.001,0.01,0.1,1} <br> max_depth={5,10,15,20,25} <br> n_estimators={50, 100, 150, 200, 250} |
| TRA | max_depth={5,10,15,20,25} <br> n_estimators={50, 100, 150, 200, 250} |

**Table 5 The Optimal Values of Hyperparameters for Machine Learning Algorithms**

| Model | Upstream prediction | On-ramp prediction | Downstream prediction |
|---|---|---|---|
| KNN | n_neighbors=11 | n_neighbors=5 | n_neighbors=9 |
| SVR | gamma=0.01<br>C=1.0 | gamma=0.01<br>C=100 | gamma=0.1<br>C=10 |
| ANN | hidden_layer_sizes=<br>(100,150) | hidden_layer_sizes=<br>(50,200) | hidden_layer_sizes=<br>(100,100) |
| XGB | learning_rate=0.1<br>max_depth=10<br>n_estimators=100 | learning_rate=0.1<br>max_depth=5<br>n_estimators=100 | learning_rate=0.1<br>max_depth=5<br>n_estimators=100 |
| TRA | max_depth=20<br>n_estimators=200 | max_depth=10<br>n_estimators=200 | max_depth=5<br>n_estimators=200 |

#### 4.2.3. Measurements of Effectiveness

Mean Absolute Error (MAE), Root Mean Square Error (RMSE), and Mean Absolute Percentage Error (MAPE) are three common criteria used to evaluate and compare prediction methods(Mahmoud et al., 2021; Yang et al., 2021). These three criteria are employed as performance measures for comparison in this study and are defined below:

$$\text{MAE} = \sqrt{\frac{1}{N}\sum_{k=1}^{N}(\hat{y}(k) - y(k))} \tag{11}$$

$$\text{RMSE} = \sqrt{\frac{1}{N}\sum_{k=1}^{N}(\hat{y}(k) - y(k))^2} \tag{12}$$

$$\text{MAPE} = \frac{1}{N}\sum_{k=1}^{N}\left|\frac{\hat{y}(k) - y(k)}{y(k)}\right| * 100\% \tag{13}$$

Where $\hat{y}(k)$ is the observed parameters at time $k$ and $y(k)$ is the corresponding predicted parameters. $N$ is the size of the testing data set (total number of time intervals).

### 4.3. Ridge Regression Results

Ridge regression was used to interpret the impacts of ramp metering control strategy changes on freeway segments. The coefficients of traffic variables related to upstream, on-ramp, and downstream segments were calculated based on known freeway segments. To choose the most influential variables, this study chose 0.5, 0.5, and 50 as the thresholds for variable filtering of speed prediction, occupancy prediction, and traffic flow prediction, respectively. That means any variable with an absolute value higher than 0.5, 0.5, and 50 would be selected as an input for transfer learning model training.

Table 6, Table 7, and Table 8 display the ridge regression results for speed prediction, occupancy prediction, and traffic flow prediction. To clarify the relationships between variables from before and after situations and reduce randomness, ridge regression was executed 10 times and the value of the coefficient for each variable was set as the average score of 10 regression results. To better identify the traffic variables, they are named the Before/After_segment position_traffic variable, following the structure. The segment positions are named up, down, and ramp, representing the upstream, downstream, and on-ramp positions.

Table 6 shows DOW, HOD, and MOH are insignificant variables that are not included in most of the cases for speed prediction. Only HOD has an impact on on-ramp speed prediction for the after situation of a ramp metering change. As expected, mean speed in the before situation is an important factor that influences mean speed prediction for an after situation. Traffic flow and traffic density play important roles in the determination of after speeds. Occupancy has trivial impacts on the prediction of the mean speed of the after situation.

**Table 6 Ridge Regression Results for Speed Prediction**

| Traffic variables | After_up_mean_speed | After_ramp_mean_speed | After_down_mean_speed |
|---|---|---|---|
| DOW | 0.01 | 0.24 | 0.01 |
| HOD | -0.06 | **-0.64** | -0.16 |
| MOH | -0.01 | 0.02 | 0.02 |
| Before_up_mean_speed | **1.98** | **0.97** | **1.53** |
| Before_ramp_mean_speed | -0.02 | **5.56** | **0.77** |
| Before_down_mean_speed | **1.67** | **-1.72** | **2.14** |
| Before_up_occupancy | -0.27 | 0.07 | -0.27 |
| Before_ramp_occupancy | -0.05 | **-0.67** | -0.15 |
| Before_up_flow | **-1.03** | **-5.48** | **-0.57** |
| Before_ramp_flow | 0.06 | **-2.91** | **-0.51** |
| Before_down_flow | **-0.76** | **8.43** | **-1.45** |
| Before_up_density | 0.30 | **5.53** | **0.51** |
| Before_ramp_density | 0.41 | **4.27** | **1.37** |
| Before_down_density | **0.97** | **-9.63** | **0.84** |

Note: Significant variables that were selected by the ridge regression models are marked with bold and underline.

Table 7 details ridge regression results for occupancy prediction. Similar to speed prediction, it shows that DOW, HOD, and MOH are insignificant variables in terms of occupancy prediction. Compared with flow rate and density, traffic speeds are influential factors in the determination of occupancy. Upstream occupancy in the before period influences both upstream and on-ramp occupancy prediction of the after period.

**Table 7 Ridge Regression Results for Occupancy Prediction**

| Traffic variables | After_up_occupancy | After_ramp_occupancy |
|---|---|---|
| DOW | 0.00 | 0.06 |
| HOD | -0.17 | -0.33 |
| MOH | -0.10 | -0.09 |
| Before_up_mean_speed | **0.56** | 0.29 |
| Before_ramp_mean_speed | 0.06 | **0.57** |
| Before_down_mean_speed | **-0.54** | **-0.64** |
| Before_up_occupancy | **3.63** | **0.59** |
| Before_ramp_occupancy | 0.07 | **4.52** |
| Before_up_flow | 0.13 | 0.06 |
| Before_ramp_flow | -0.01 | **0.73** |
| Before_down_flow | **-0.51** | **-0.87** |
| Before_up_density | **0.55** | 0.33 |
| Before_ramp_density | 0.10 | -0.08 |
| Before_down_density | 0.02 | 0.11 |

Note: Significant variables that were selected by the ridge regression models are marked with bold and underline.

Table 8 presents ridge regression results for traffic flow prediction. It turns out that DOW, HOD, and MOH are insignificant variables in terms of flow rate prediction. Upstream speed in the before period has negative associations with both upstream and downstream speeds in the after period. On the contrary, downstream speed in the before period has positive associations with both upstream and downstream speeds in the after period. Compared with on-ramp and downstream traffic flow rates in the before period, the upstream traffic flow rate in the before period can strongly influence traffic flow prediction of all segments in the after periods. It is apparent that traffic speeds, flow rates, and densities in the before period can heavily impact traffic flow prediction of the after period compared with occupancies in the before period.

**Table 8 Ridge Regression Results for Traffic Flow Prediction**

| Traffic variables | After_up_flow | After_ramp_flow | After_down_flow |
|---|---|---|---|
| DOW | -9.14 | -3.66 | -7.10 |
| HOD | -23.71 | -5.48 | -24.65 |
| MOH | -14.04 | -2.70 | -14.07 |
| Before_up_mean_speed | **-75.23** | 33.89 | **-75.85** |
| Before_ramp_mean_speed | -21.51 | 6.80 | -9.10 |
| Before_down_mean_speed | **100.07** | -25.15 | **90.62** |
| Before_up_occupancy | -13.05 | 2.64 | -8.40 |
| Before_ramp_occupancy | 2.58 | -5.06 | -3.55 |
| Before_up_flow | **517.23** | **-89.03** | **215.98** |
| Before_ramp_flow | 40.57 | **117.34** | 32.77 |
| Before_down_flow | **-257.76** | 34.67 | 41.87 |
| Before_up_density | **-273.67** | **103.06** | **-261.02** |
| Before_ramp_density | **-83.42** | 7.32 | **-61.61** |
| Before_down_density | **315.56** | **-50.65** | **287.77** |

Note: Significant variables that were selected by the ridge regression models are marked with bold and underline.

In summary, the inclusion of traffic variables coming from upstream, downstream, and on-ramp ensures that spatial information was considered in the prediction process. Under different prediction cases, the variables that have the most significant impacts on the prediction results vary case by case. In comparison to occupancy prediction and flow rate prediction, speed prediction is influenced by more variables. Overall, ridge regression ranks the variables that affect the prediction results of after period, which provides an approach for better understanding the interaction between freeway traffic and ramp metering control strategy.

### 4.4. Prediction Results

The actual inputs for the transfer learning model training process only contained the traffic variables selected by ridge regression models. The outputs of transfer learning models included upstream mean speeds, on-ramp mean speeds, downstream mean speeds, upstream occupancies, on-ramp occupancies, upstream flow rates, on-ramp flow rates, and downstream flow rates of all study sites. This section shows the prediction performance comparison between different models. All the results presented were the average values of ten runs of the experiments.

#### 4.4.1. Speed Prediction Results

Table 9, Figure 7, and Figure 8 summarize the speed prediction performance comparison between four state-of-the-art models (KNN, SVR, ANN, and XGB) and the proposed TRA model in terms of MAE, RMSE, and MAPE. The proposed TRA models acquire the highest prediction accuracy in most cases in terms of MAE. SVR achieves the best prediction performance for downstream speed prediction of #201, XGB outperforms other models when predicting downstream speed for #130, ANN has the lowest MAE when predicting on-ramp speed for #312. Within all prediction scenarios for the proposed method, the MAEs range from 0.47 mph to 3.86 mph, RMSEs range from 0.61 to 5.07, and MAPEs range from 0.67% to 10.37%. As shown in Figure 7 and Figure 8, compared to upstream and downstream speed prediction, all the models have the lowest prediction accuracy for on-ramp speed prediction. As on-ramps are heavily influenced by ramp metering, which produces more uncertainty and variation on traffic speed, it is understandable that it would be difficult for machine learning models to predict on-ramp speed with high accuracy.

**Table 9 MAE Comparison between Different Models for Speed Prediction**

| Location | Model | Loop detector numbering | | | | | | |
|---|---|---|---|---|---|---|---|---|
| | | #130 | #190 | #193 | #195 | #198 | #201 | #294 |
| Upstream | KNN | 2.89 | 2.05 | 2.96 | 1.75 | 2.35 | 3.39 | 1.11 |
| | SVR | 1.94 | 1.81 | 3.12 | 1.84 | 1.61 | 2.87 | 1.52 |
| | ANN | 2.33 | 1.51 | 2.82 | 2.07 | 2.29 | 4.17 | 1.33 |
| | XGB | 2.79 | 1.46 | 2.21 | 1.78 | 1.47 | 3.04 | 1.33 |
| | TRA | **1.78** | **1.09** | **1.55** | **1.29** | **1.30** | **1.94** | **0.91** |
| On-ramp | KNN | 3.31 | 2.64 | 5.76 | 3.86 | 4.52 | 3.42 | 4.09 |
| | SVR | 5.93 | 2.34 | 6.72 | 4.07 | 6.34 | 3.00 | 5.07 |
| | ANN | 4.37 | 3.10 | 7.28 | 4.75 | 5.81 | 3.13 | 4.33 |
| | XGB | 4.36 | 2.41 | 5.56 | 3.87 | 6.21 | 3.65 | 3.77 |
| | TRA | **1.62** | **1.92** | **3.86** | **2.67** | **3.83** | **2.51** | **2.04** |
| Downstream | KNN | 2.24 | 2.02 | 1.72 | 2.56 | 2.69 | 1.91 | 0.87 |
| | SVR | 1.31 | 1.85 | 2.18 | 2.42 | 2.63 | **1.38** | 0.94 |
| | ANN | 2.80 | 1.77 | 1.98 | 2.31 | 2.76 | 2.00 | 1.56 |
| | XGB | **1.28** | 1.55 | 1.59 | 2.02 | 2.92 | 1.62 | 1.16 |
| | TRA | 1.30 | **1.35** | **1.46** | **1.75** | **1.83** | 1.58 | **0.83** |
| Location | Model | Loop detector numbering | | | | | | |
| | | #299 | #300 | #303 | #306 | #309 | #312 | #622 |
| Upstream | KNN | 0.92 | 2.08 | 1.07 | 1.01 | 0.91 | 2.11 | 2.30 |
| | SVR | 0.87 | 1.53 | 1.08 | 1.25 | 1.55 | 2.69 | 2.47 |
|  | ANN | 0.96 | 1.72 | 1.70 | 1.39 | 1.94 | 3.01 | 1.37 |
|  | XGB | 0.95 | 1.53 | 1.03 | 1.14 | 0.95 | 2.14 | 3.14 |
|  | TRA | **0.50** | **1.31** | **0.95** | **0.82** | **0.59** | **1.35** | **1.24** |
|  | KNN | 3.53 | 5.77 | 3.10 | 3.61 | 5.50 | 4.38 | 6.40 |
|  | SVR | 3.59 | 5.87 | 3.09 | 2.79 | 5.54 | 2.14 | 4.38 |
| On-ramp | ANN | 3.97 | 5.87 | 2.47 | 3.61 | 5.65 | **1.75** | 2.89 |
|  | XGB | 2.75 | 3.44 | 2.59 | 2.56 | 4.89 | 2.61 | 5.88 |
|  | TRA | **1.48** | **2.01** | **2.34** | **2.19** | **3.25** | 2.03 | **1.63** |
|  | KNN | 1.82 | 1.27 | 0.84 | 1.42 | 0.98 | 0.62 | 2.52 |
|  | SVR | 2.79 | 0.82 | 0.77 | 1.51 | 1.28 | 0.78 | 3.68 |
| Downstream | ANN | 2.77 | 0.91 | 0.89 | 1.98 | 1.57 | 0.98 | 2.40 |
|  | XGB | 2.44 | 1.07 | 1.00 | 1.56 | 1.09 | 0.84 | 4.12 |
|  | TRA | **0.90** | **0.59** | **0.82** | **0.98** | **0.79** | **0.47** | **0.92** |

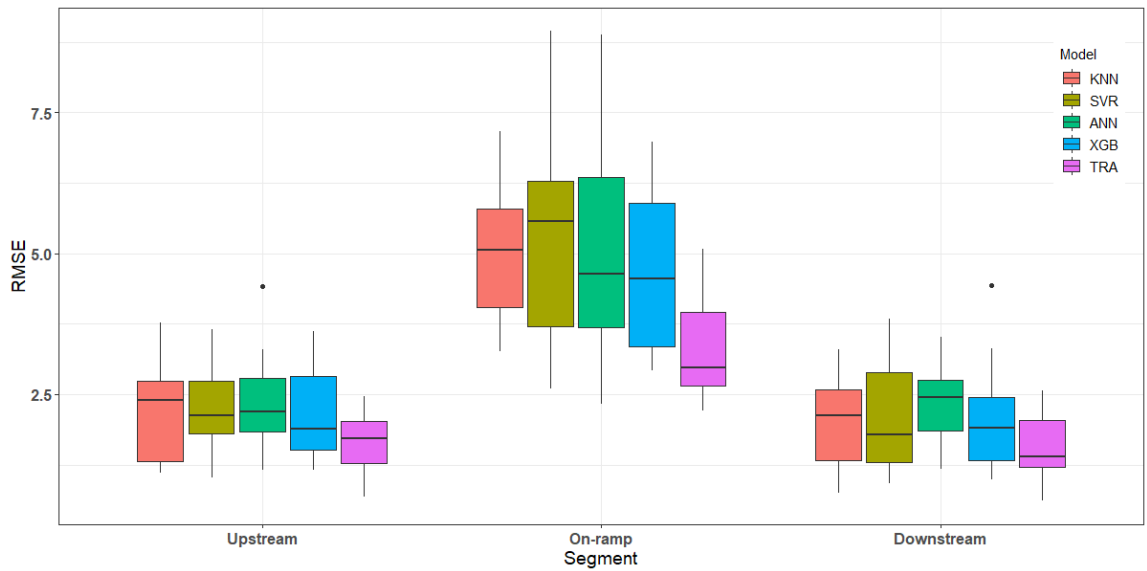

**Figure 7 RMSE Comparison between Different Models for Speed Prediction**

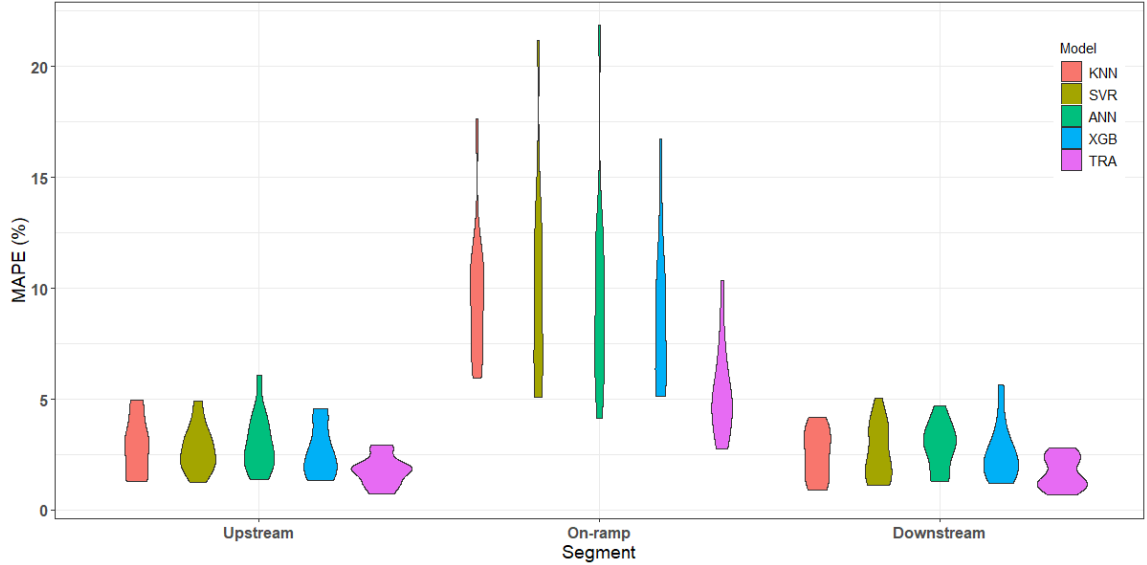

**Figure 8 MAPE Comparison between Different Models for Speed Prediction**

### 4.4.2. Occupancy Prediction Results

Table 10, Figure 9, and Figure 10 compare the occupancy prediction performance between four state-of-the-art models (KNN, SVR, ANN, and XGB) and the proposed TRA model in terms of MAE, RMSE, and MAPE. The proposed TRA models acquire the highest prediction accuracy in most cases in terms of MAE except upstream occupancy prediction of #190. Within all prediction scenarios for the proposed method, the MAEs range from 0.35% to 2.58%, RMSEs range from 0.57 to 3.56, and MAPEs range from 4.77% to 32.87%. Similar to speed prediction, compared with upstream occupancy prediction, all the models have the lowest prediction accuracy for on-ramp occupancy prediction. This can also be explained by the influence of ramp metering which produces more uncertainty and variation on occupancy.

**Table 10 MAE Comparison between Different Models for Occupancy Prediction**

| Location | Model | Loop detector numbering | | | | | | |
|---|---|---|---|---|---|---|---|---|
| | | #130 | #190 | #193 | #195 | #198 | #201 | #294 |
| Upstream | KNN | 1.63 | 1.95 | 1.88 | 1.47 | 1.71 | 1.56 | 1.26 |
| | SVR | 1.33 | **1.83** | 1.42 | 1.04 | 1.45 | 1.37 | 1.12 |
| | ANN | 1.45 | 2.18 | 2.43 | 1.40 | 1.91 | 1.44 | 1.39 |
| | XGB | 1.39 | 2.24 | 2.47 | 1.15 | 1.79 | 1.40 | 1.33 |
| | TRA | **1.03** | 2.02 | **0.88** | **0.91** | **1.23** | **0.97** | **1.00** |
| On-ramp | KNN | 0.81 | 3.44 | 1.67 | 5.09 | 2.72 | 1.53 | 2.76 |
| | SVR | 0.76 | 2.60 | 1.35 | 5.61 | 2.31 | 1.02 | 2.08 |
| | ANN | 0.82 | 2.51 | 1.51 | 4.67 | 2.20 | 1.70 | 2.80 |
| | XGB | 0.66 | 2.77 | 1.47 | 7.27 | 1.79 | 1.19 | 3.33 |
| | TRA | **0.49** | **2.40** | **0.89** | **2.58** | **1.50** | **0.51** | **1.97** |
| Location | Model | Loop detector numbering | | | | | | |
| | | #299 | #300 | #303 | #306 | #309 | #312 | #622 |
| Upstream | KNN | 0.51 | 0.70 | 0.54 | 0.58 | 1.70 | 0.51 | 0.72 |
| | SVR | 0.46 | 0.62 | 0.54 | 0.56 | 1.18 | 0.42 | 0.42 |

|         | ANN | 0.48 | 0.76 | 1.11 | 0.65 | 1.21 | 0.50 | 0.46 |
|         | XGB | 0.60 | 0.85 | 0.83 | 0.57 | 0.83 | 0.67 | 0.51 |
|         | TRA | **0.44** | **0.51** | **0.35** | **0.52** | **0.45** | **0.40** | **0.36** |
|         | KNN | 1.10 | 1.01 | 1.10 | 0.51 | 0.87 | 1.08 | 1.95 |
|         | SVR | 0.59 | 1.07 | 0.78 | 0.64 | 1.32 | 0.77 | 1.66 |
| On-ramp | ANN | 1.01 | 1.16 | 1.20 | 0.53 | 0.95 | 0.96 | 2.02 |
|         | XGB | 0.78 | 1.01 | 0.79 | 0.54 | 1.55 | 1.41 | 1.56 |
|         | TRA | **0.43** | **0.95** | **0.48** | **0.41** | **0.60** | **0.47** | **0.92** |

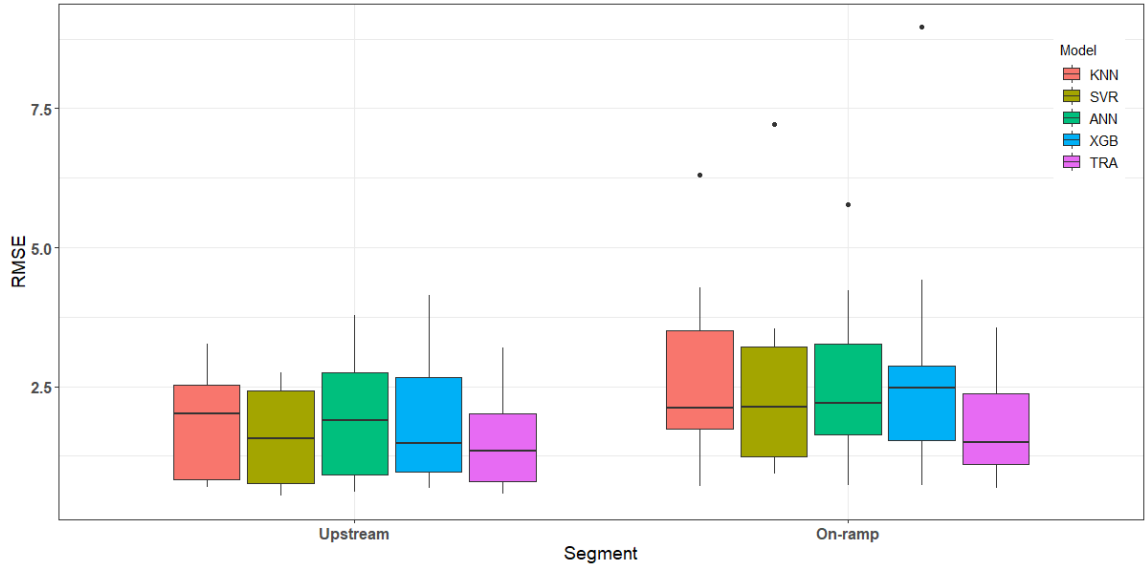

**Figure 9 RMSE Comparison between Different Models for Occupancy Prediction**

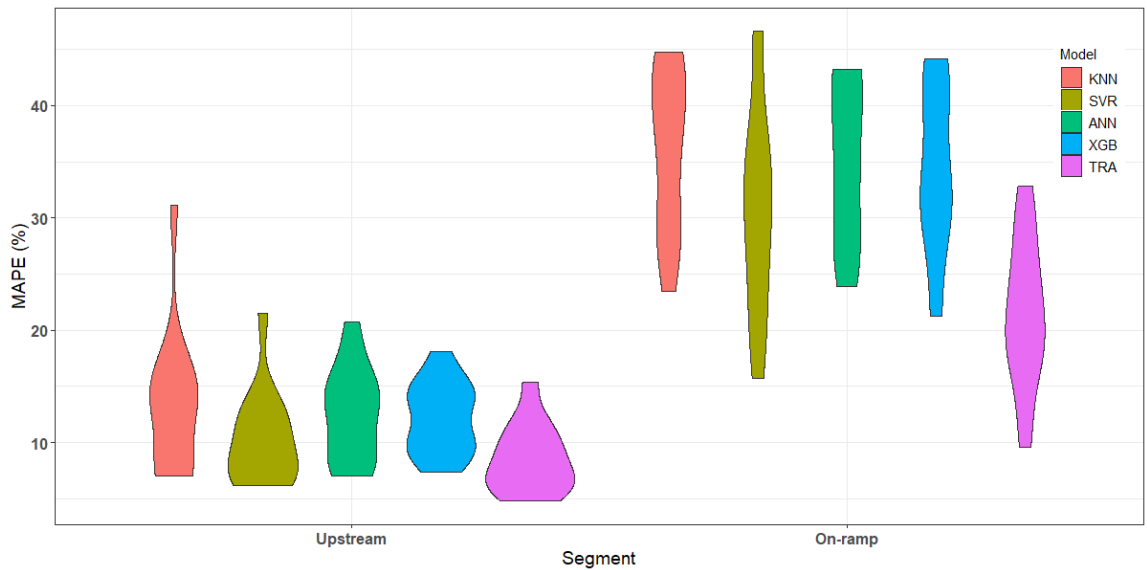

**Figure 10 MAPE Comparison between Different Models for Occupancy Prediction**

### 4.4.3. Flow Rate Prediction Results

Table 11, Figure 11, and Figure 12 illustrate the flow rate prediction performance comparison between four state-of-the-art models (KNN, SVR, ANN, and XGB) and the proposed TRA model in terms of MAE, RMSE, and MAPE. KNN, SVR, and ANN beat the proposed model in certain cases, but overall, the proposed TRA models obtain the highest prediction accuracy in most cases. Within all prediction scenarios for the proposed method, the MAEs range from 17.49 veh/hr/ln to 235.02 veh/hr/ln, RMSEs range from 23.35 to 282.66, and MAPEs range from 3.24% to 22.95%. It can be seen that in Figure 11, TRA models have relatively low variation in terms of RMSE compared with other models. Figure 12 reveals that for both upstream and downstream flow rate prediction, the majority of prediction results produced by TRA models are less than 5.00% MAPE.

**Table 11 MAE Comparison between Different Models for Flow Rate Prediction**

| Location | Model | Loop detector numbering | | | | | | |
|---|---|---|---|---|---|---|---|---|
| | | #130 | #190 | #193 | #195 | #198 | #201 | #294 |
| Upstream | KNN | 99.63 | 106.60 | 76.22 | 92.40 | **207.40** | 63.90 | 55.44 |
| | SVR | 87.73 | 89.53 | 96.38 | 82.15 | 231.37 | 73.22 | 58.37 |
| | ANN | 72.48 | 98.28 | 99.40 | 119.98 | 343.04 | 55.12 | 50.80 |
| | XGB | 78.20 | 136.42 | 64.90 | 68.40 | 271.64 | 64.25 | 52.48 |
| | TRA | **57.80** | **58.08** | **47.57** | **50.88** | 235.02 | **53.43** | **36.91** |
| On-ramp | KNN | 48.63 | 29.59 | 34.96 | 34.13 | 64.30 | 21.27 | 44.88 |
| | SVR | 38.43 | 24.55 | 25.67 | 27.14 | 59.81 | 18.79 | 39.28 |
| | ANN | 39.26 | **23.13** | 24.71 | 25.57 | 33.50 | 18.74 | 35.24 |
| | XGB | 38.77 | 31.41 | 23.93 | 32.37 | 50.03 | 22.82 | 39.56 |
| | TRA | **37.62** | 23.48 | **17.90** | **25.50** | **31.17** | **18.14** | **31.80** |
| Downstream | KNN | 95.91 | 91.38 | 124.56 | 98.84 | 270.66 | 83.53 | 66.74 |
| | SVR | 72.97 | 78.41 | 189.22 | 59.57 | 258.89 | 92.67 | 51.66 |
| | ANN | 74.66 | **62.66** | 101.65 | 108.97 | 263.21 | 62.22 | 50.1 |
| | XGB | 88.63 | 120.56 | 93.81 | 94.71 | 212.98 | 68.24 | 51.58 |
| | TRA | **67.2** | 74.08 | **56.74** | **48.94** | **186.43** | **56.42** | **43.17** |
| Location | Model | Loop detector numbering | | | | | | |
| | | #299 | #300 | #303 | #306 | #309 | #312 | #622 |
| Upstream | KNN | 50.88 | 106.04 | 78.85 | 60.98 | 59.27 | 58.70 | 81.37 |
| | SVR | 48.81 | 106.30 | 89.86 | 75.07 | 53.08 | 52.52 | 192.34 |
| | ANN | 61.43 | 91.24 | 56.38 | 50.96 | **43.82** | 62.72 | 52.94 |
| | XGB | 73.87 | 102.70 | 90.55 | 78.00 | 75.20 | 48.78 | 50.26 |
| | TRA | **37.71** | **62.55** | **51.16** | **48.45** | 44.09 | **37.96** | **38.27** |
| On-ramp | KNN | 61.57 | 51.00 | 39.60 | 36.77 | 35.85 | 23.66 | 40.12 |
| | SVR | 50.86 | 48.25 | **24.37** | **19.88** | 32.01 | 24.93 | 26.10 |
| | ANN | 32.20 | 48.38 | 26.87 | 26.29 | 33.34 | 22.46 | 21.45 |
| | XGB | 34.10 | 53.29 | 36.91 | 31.45 | 37.76 | 28.92 | 25.61 |
| | TRA | **17.49** | **41.14** | 25.37 | 22.73 | **22.74** | **20.18** | **19.31** |
| Downstream | KNN | 45.67 | 103.33 | 58.45 | 55.86 | 146.46 | 81.44 | 79.54 |
| | SVR | 61.91 | 100.46 | 55.23 | **47.08** | 126.05 | 54.14 | 93.12 |
| | ANN | 58.3 | 100.35 | 50.33 | 61.64 | 56.37 | 39.04 | 47.18 |
| | XGB | 55.16 | 108.93 | 92.29 | 50.36 | 113.84 | 42.85 | 83.02 |

| | TRA | **36.38** | **75.19** | **40.66** | 50.56 | **43.17** | 33.16 | **36.76** |

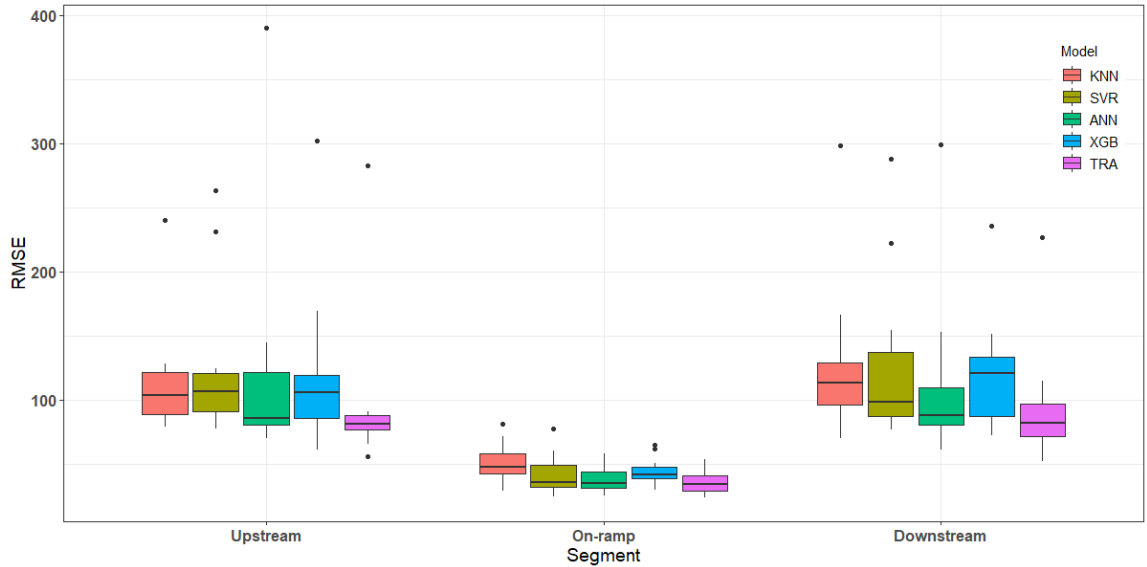

**Figure 11 RMSE Comparison between Different Models for Flow Rate Prediction**

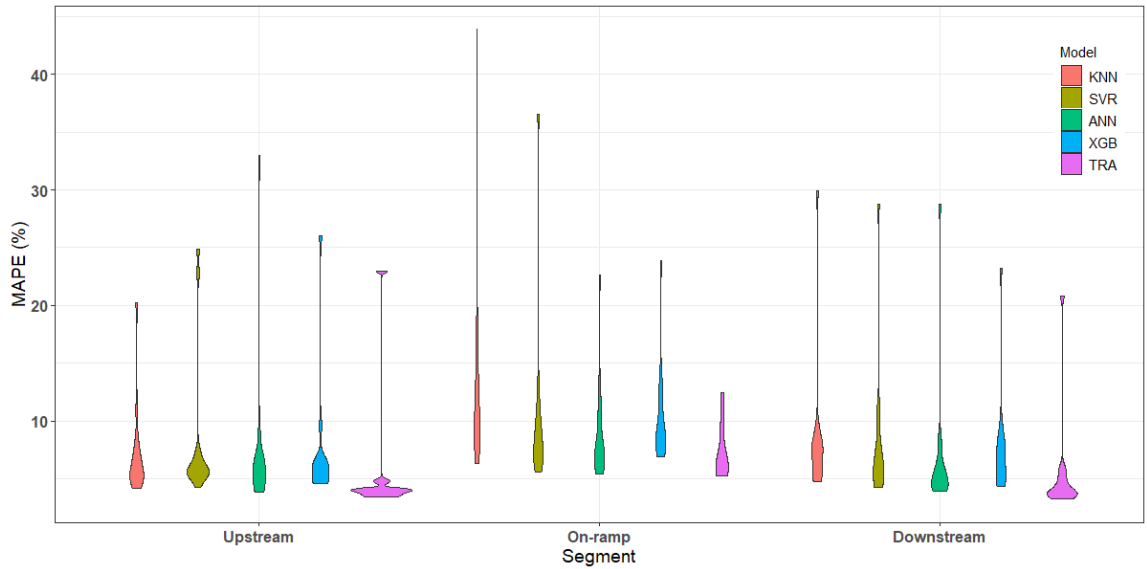

**Figure 12 MAPE Comparison between Different Models for Flow Rate Prediction**

Compared with other state-of-the-art machine learning models, TRA has the highest accuracy when predicting speed, occupancy, and flow rate. This is intuitive, as TRA relaxes the assumption that the data underlying distributions of the source and target domains must be the same. Therefore, it can handle the prediction of speed, occupancy, and flow rate in different freeway segments with different traffic patterns, distributions, and characteristics. Thus, the proposed TRA model can develop scene-specific models which provide superior prediction accuracy compared to generic models.

## 5. CONCLUSION

In this study, an innovative framework is developed for predicting freeway traffic parameters when a ramp metering control strategy is altered. The predicted parameters can then be used for assisting in ramp metering evaluation. In this framework, the spatial-temporal traffic features from before and after situations are temporally corrected and mapped in chronological order to allow the application of ridge regression, in order to easily provide insights about the association between traffic variables of before and after situations. After selecting the most influential traffic variables through ridge regression, the transfer learning model Two-stage TrAdaBoost.R2 is used to predict freeway traffic parameters. The proposed method exhibits superior spatial-temporal transferability for new freeway segments in comparison to traditional machine learning methods, as shown in its evaluation by freeway traffic data collected from the southbound stretch of SR-51 in the Phoenix Metropolitan area. Engineers may use the suggested approach to predict traffic parameters for new freeway segments, facilitating early identification of potential issues. By comparing predicted traffic parameters with those in the pre-implementation period, the impact of ramp metering control strategies can be evaluated. The findings, such as changes in traffic speed, can provide valuable insights into the feasibility of applying ramp metering control strategies to on-ramps. This information can help engineers tailor ramp metering control strategies to specific on-ramps and fine-tune their designs before actual deployment. As a common issue with data-driven research, lack of data will be a limitation in implementing such methodology. Future implementation should consider data sufficiency before employing the proposed method.

As the framework has high flexibility allowing for the incorporation of multiple variables, future work could consider adding more temporal and spatial traffic features to the proposed model. Also, weather conditions, incident events, and other factors that might impact the traffic operation can be included in the model to further improve the generalization abilities of the model. This paper attempts to learn the influence of ramp metering on known freeway segments and predict the impact of ramp metering on new segments using the knowledge learned. In the future, more advanced machine learning methods and more sophisticated input features could be applied to further improve the prediction performance. The predicted speed, traffic flow, and occupancy could be used to infer ramp queue spill and other ramp metering-related issues. Future research could further investigate the possibility of using predicted traffic parameters for the proactive management of ramp metering-related issues.

**ACKNOWLEDGMENTS**

The authors would like to acknowledge the Arizona Department of Transportation for graciously sharing the crowdsourced and loop detector data that made this paper possible. Special thanks to Lilly Cottam and Wei Xu for their assistance in English proofreading.